\documentclass[11pt,conference]{IEEEtran}
\usepackage{graphicx} 
\usepackage{fancyvrb}
\usepackage{listings}
\usepackage{caption}
\usepackage{color}
\usepackage{multirow}
\usepackage{booktabs}
\usepackage{hyperref}

\usepackage{tikz}
\usepackage{pgfplots}
\usepgfmodule{oo}
\usepackage{ifthen}
\usepackage{etoolbox}
\usepackage[nomessages]{fp}

\usepackage{tikz}
\usepgfmodule{oo}
\usepackage{ifthen}
\usepackage{etoolbox}
\usepackage[nomessages]{fp}

\usepackage{pgf} 
\usepackage{fp}

\pgfmathsetseed{\number\pdfrandomseed}

\def\splitlist#1,#2\relax{\def\firstid{#1}\def\secondid{#2}}

\newcommand{\randfloat}[1]{%
 \pgfmathsetmacro{\thenum}{random(0,100000)}
 \FPeval{\result}{clip((\thenum-50000)/50000)}
 \expandafter\let\csname #1\endcsname\result 
}%

\newcommand{\boundrandfloat}[3]{%
    \FPrandom{\result}%
    \FPsub{\range}{#3}{#2}%
    \FPmul{\scaled}{\result}{\range}%
    \FPadd{#1}{\scaled}{#2}%
}

\makeatletter
\protected\def\zip{%
  \begingroup
  \@ifstar{\def\cnta{1}\@zip}
    {\def\cnta{0}\@zip}%
}
\def\@zip#1#2#3#4{%
  \def\tempa##1##2{%
    \edef##2{%
      \ifnum\cnta=\@ne\else\expandafter\@firstoftwo\fi
      \unexpanded\expandafter{##1}%
    }%
  }%
  \tempa{#2}\tempb\tempa{#3}\tempa
  \def\cnta{0}\def#4{}%
  \foreach \x in \tempb{%
    \xdef\cnta{\the\numexpr\cnta+1}%
    \gdef\cntb{0}%
    \foreach \y in \tempa{%
      \xdef\cntb{\the\numexpr\cntb+1}%
      \ifnum\cntb=\cnta\relax
        \xdef#4{#4\ifx#4\empty\else,\fi\x#1\y}%
        \breakforeach
      \fi
    }%
  }%
  \endgroup
}
\makeatother

\makeatletter
\newcommand{\enqueue}[2]{%
    \@ifundefined{#1@tail}{%
        \expandafter\gdef\csname #1@tail\endcsname{0}%
        \expandafter\gdef\csname #1@head\endcsname{0}%
    }{}%
    \edef\@temp{\the\numexpr\csname #1@tail\endcsname+1\relax}%
    \expandafter\xdef\csname #1@\@temp\endcsname{#2}%
    \expandafter\xdef\csname #1@tail\endcsname{\@temp}%
}
\newcommand{\dequeue}[2]{%
    \@ifundefined{#1@head}{%
        Queue is empty
    }{%
        \ifnum\csname #1@head\endcsname=\csname #1@tail\endcsname
            Queue is empty
        \else
            \edef\@temp{\the\numexpr\csname #1@head\endcsname+1\relax}%
            \expandafter\let\expandafter#2\expandafter=\csname #1@\@temp\endcsname%
            \expandafter\global\expandafter\let\csname #1@\@temp\endcsname\relax%
            \expandafter\xdef\csname #1@head\endcsname{\@temp}%
        \fi
    }%
}
\newcommand{\initQueue}[1]{%
    \global\expandafter\let\csname #1@tail\endcsname\relax%
    \global\expandafter\let\csname #1@head\endcsname\relax%
}
\makeatother

\newcommand{\peekQueue}[1]{%
    \ifnum\csname #1@head\endcsname=\csname #1@tail\endcsname
        Queue is empty
    \else
        \def\tempHead{\csname #1@head\endcsname}
        \def\tempTail{\csname #1@tail\endcsname}
        \def\tempItem{}
        {\loop
            \edef\tempItem{\csname #1@\tempHead\endcsname}
            \tempItem
            \ifnum\tempHead<\tempTail
                \edef\tempHead{\the\numexpr\tempHead+1\relax}
                \repeat}
    \fi
}

\usepackage{xstring}

\def\splicelist#1{
\StrCount{#1}{,}[\numofelem]
\ifnum\numofelem>0\relax
    \StrBefore[1]{#1}{,}[\myhead]%
    \StrBehind[1]{#1}{,}[\mytail]%
    \StrBehind[\numofelem]{#1}{,}[\mylast]%
\else
    \let\myhead#1%
    \let\mylast#1%
    \def\mytail{N/A}
\fi
}

\makeatletter
\def\splitstringslash#1/#2\relax{%
    \def\inputID{#1}%
    \def\weightID{#2}%
}

\makeatother

\newcommand{\listlength}[2]{%
    \IfStrEq{#2}{}{%
        \edef#1{0}%
    }{%
        \StrCount{#2}{,}[\numofcommas]%
        \edef#1{\number\numexpr\numofcommas+1\relax}%
    }%
}

\newboolean{debug} 
\newcommand{\debug}[2]{%
    \ifthenelse{\boolean{debug}}{%
        \ifthenelse{\equal{#2}{LOW}}{%
            \message{DEBUG: #1 ^^J}%
        }{}%
        \ifthenelse{\equal{#2}{MED}}{%
            \ifthenelse{\equal{\debuglevel}{MED}\OR\equal{\debuglevel}{HIGH}}{%
                \message{DEBUG: #1 ^^J}%
            }{}%
        }{}%
        \ifthenelse{\equal{#2}{HIGH}}{%
            \ifthenelse{\equal{\debuglevel}{HIGH}}{%
                \message{DEBUG: #1 ^^J}%
            }{}%
        }{}%
    }{}%
}

\pgfooclass{Value}{
    \attribute data;
    \attribute grad;
    \attribute prev; 
    \attribute next; 
    \attribute op; 
    \attribute isparam; 
    \attribute gradcounter; 
    \attribute donebackwards; 
    \attribute computegraph; 
    
    \method Value(#1,#2,#3,#4) {
        \pgfooeset{data}{#1}
        \pgfooeset{grad}{0.0}
        \pgfooeset{prev}{#2}
        \pgfooeset{next}{}
        \pgfooeset{op}{#3}
        \pgfooeset{isparam}{#4}
        \pgfooeset{gradcounter}{0.0}
        \pgfooeset{donebackwards}{0.0}
        \pgfooeset{computegraph}{}
    }

    \method getdata(#1) {
        \pgfooget{data}{#1}
    }

    \method getgrad(#1) {
        \pgfooget{grad}{#1}
    }

    \method getprev(#1) {
        \pgfooget{prev}{#1}
    }

    \method getnext(#1) {
        \pgfooget{next}{#1}
    }

    \method getop(#1) {
        \pgfooget{op}{#1}
    }

    \method getisparam(#1) {
        \pgfooget{isparam}{#1}
    }

    \method getgradcounter(#1) {
        \pgfooget{gradcounter}{#1}
    }

    \method getdonebackwards(#1) {
        \pgfooget{donebackwards}{#1}
    }

    \method getcomputegraph(#1) {
        \pgfooget{computegraph}{#1}
    }

    \method setdata(#1) {
        \pgfooeset{data}{#1}
    }
    
    \method setgrad(#1) {
        \pgfooeset{grad}{#1}
    }
    
    \method setprev(#1) {
        \pgfooeset{prev}{#1}
    }
    
    \method setop(#1) {
        \pgfooeset{op}{#1}
    }

    \method appendtonext(#1) {
        \pgfooget{next}{\nextIDs}
        \ifx\nextIDs\empty
            \pgfooeset{next}{#1}
        \else
            \edef\nextIDs{\nextIDs,#1}
            \pgfooeset{next}{\nextIDs}
        \fi
    }

    \method appendtocomputegraph(#1) {
        \pgfooget{computegraph}{\tempcomputegraph}
        \ifx\tempcomputegraph\empty
            \pgfooeset{computegraph}{#1}
        \else
            \edef\tempcomputegraph{\tempcomputegraph,#1}
            \pgfooeset{computegraph}{\tempcomputegraph}
        \fi
    }

    \method setgradcounter(#1) {
        \pgfooeset{gradcounter}{#1}
    }

    \method setdonebackwards(#1) {
        \pgfooeset{donebackwards}{#1}
    }

    \method increasegradcounter() {
        \pgfooget{gradcounter}{\currentgradcounter}
        \FPeval\newgradcounter{(\currentgradcounter) + (1)}
        \pgfooeset{gradcounter}{\newgradcounter}
    }

    \method show() {
        \pgfoothis.get id(\tempid)
        
        \pgfooget{data}{\tempdata}
        \pgfooget{grad}{\tempgrad}
        \pgfooget{prev}{\tempprev}
        \pgfooget{next}{\tempnext}
        \pgfooget{op}{\tempop}
        \pgfooget{isparam}{\tempparam}
        \pgfooget{gradcounter}{\tempgradcounter}
        Value(self: \tempid, data: \tempdata, grad: \tempgrad, prev: \tempprev, next: \tempnext, op: \tempop, isparam: \tempparam, GC: \tempgradcounter)
    }

    \method debugmsg(#1) {
        \pgfoothis.get id(\tempid)
        
        \pgfooget{data}{\tempdata}
        \pgfooget{grad}{\tempgrad}
        \pgfooget{prev}{\tempprev}
        \pgfooget{next}{\tempnext}
        \pgfooget{op}{\tempop}
        \pgfooget{isparam}{\tempparam}
        \pgfooget{gradcounter}{\tempgradcounter}
        \debug{Value(self: \tempid, data: \tempdata, grad: \tempgrad, prev: \tempprev, next: \tempnext, op: \tempop, isparam: \tempparam, GC: \tempgradcounter)}{#1}
    }
    
    \method add(#1,#2) { 
        \pgfooget{data}{\selfdata}
        #1.getdata(\otherdata)
        
        \FPeval\newdata{(\selfdata) + (\otherdata)}

        \pgfoothis.get id(\selfid)
        #1.get id(\otherid)

        \pgfooget{next}{\nextIDs}
        \ifx\nextIDs\empty
            \debug{Value: \selfid\space - Inside Add, First Forward Pass}{HIGH}
            \expanded{%
              \noexpand\pgfoonew\expandafter\noexpand\csname #2\endcsname=new Value(\newdata,{\selfid,\otherid},+,0)%
            }%
            \csname #2\endcsname.get id(\nextID)
            \pgfooeset{next}{\nextID}
            \pgfooobj{\otherid}.appendtonext(\nextID)
        \else
            \debug{Value: \selfid\space - Inside Add, Subsequent Forward Pass}{HIGH}
            \edef\notCreated{1}
            \splicelist{\nextIDs}
            \listlength{\numNextIDs}{\nextIDs}

            \pgfplotsforeachungrouped \i in {1,...,\numNextIDs} {
                \ifnum\notCreated=0
                \else
                    \edef\addNextHead{\myhead}
                    \edef\addNextTail{\mytail}

                    \pgfooobj{\addNextHead}.getprev(\tempParents)
                    \expandafter\splitlist\tempParents\relax 
                    \debug{Value: \selfid\space - Inside Add, Checking \otherid\space matches \secondid}{HIGH}
    
                    \ifx\otherid\secondid
                        \edef\notCreated{0}
                        \pgfooobj{\addNextHead}.setdata(\newdata)
                        \debug{Value: \selfid\space - Inside Add, Updating data of Value ID: \addNextHead}{HIGH}
                        \pgfooobj{\addNextHead}.get handle(\childobj)
                        \expandafter\let\csname #2\endcsname = \childobj
                    \fi
                \fi
                \splicelist{\addNextTail}
            }

            \ifnum\notCreated=1
                \debug{Value: \selfid\space - Inside Add, New inputs so creating new Value}{HIGH}
                \expanded{%
                  \noexpand\pgfoonew\expandafter\noexpand\csname #2\endcsname=new Value(\newdata,{\selfid,\otherid},+,0)%
                }%
                \csname #2\endcsname.get id(\nextID)
                \pgfoothis.appendtonext(\nextID)
                \pgfooobj{\otherid}.appendtonext(\nextID)
            \fi
        \fi
    }

    \method subtract(#1,#2) { 
        \pgfooget{data}{\selfdata}
        #1.getdata(\otherdata)
        \FPeval\newdata{(\selfdata) - (\otherdata)}

        \pgfoothis.get id(\selfid)
        #1.get id(\otherid)
        
        \pgfooget{child}{\selfchild}
        \ifnum\selfchild=0
            \expanded{%
              \noexpand\pgfoonew\expandafter\noexpand\csname #2\endcsname=new Value(\newdata,{\selfid,\otherid},+,0)%
            }%
            \csname #2\endcsname.get id(\childid)
            \pgfooeset{child}{\childid}
        \else
            \pgfooobj{\selfchild}.setdata(\newdata)
            \pgfooobj{\selfchild}.get handle(\childobj)
            \expandafter\let\csname #2\endcsname = \childobj
        \fi
    }

    \method multiply(#1,#2) { 
        \pgfooget{data}{\selfdata}
        #1.getdata(\otherdata)
        \FPeval\newdata{(\selfdata) * (\otherdata)}

        \pgfoothis.get id(\selfid)
        #1.get id(\otherid)

        \pgfooget{next}{\nextIDs}
        \ifx\nextIDs\empty
            \expanded{%
              \noexpand\pgfoonew\expandafter\noexpand\csname #2\endcsname=new Value(\newdata,{\selfid,\otherid},*,0)%
            }%
            \csname #2\endcsname.get id(\nextID)
            \pgfooeset{next}{\nextID}
            \pgfooobj{\otherid}.appendtonext(\nextID)
        \else
            \edef\notCreated{1}
            \splicelist{\nextIDs}
            \listlength{\numNextIDs}{\nextIDs}

            \pgfplotsforeachungrouped \i in {1,...,\numNextIDs} {
                \ifnum\notCreated=0
                \else
                    \edef\mulNextHead{\myhead}
                    \edef\mulNextTail{\mytail}
                    \pgfooobj{\mulNextHead}.getprev(\tempParents)
                    \expandafter\splitlist\tempParents\relax 
    
                    \ifx\otherid\secondid
                        \edef\notCreated{0}
                        \pgfooobj{\mulNextHead}.setdata(\newdata)
                        \debug{Updating object with ID \mulNextHead}{HIGH}
                        \pgfooobj{\mulNextHead}.get handle(\childobj)
                        \expandafter\let\csname #2\endcsname = \childobj
                    \fi
                \fi
                \splicelist{\mulNextTail}
            }

            \ifnum\notCreated=1
                \expanded{%
                  \noexpand\pgfoonew\expandafter\noexpand\csname #2\endcsname=new Value(\newdata,{\selfid,\otherid},*,0)%
                }%
                \csname #2\endcsname.get id(\nextID)
                \pgfoothis.appendtonext(\nextID)
                \pgfooobj{\otherid}.appendtonext(\nextID)
            \fi
        \fi
    }

    \method divide(#1,#2) { 
        \pgfooget{data}{\selfdata}
        #1.getdata(\otherdata)
        \FPeval\newdata{(\selfdata) / (\otherdata)}

        \pgfoothis.get id(\selfid)
        #1.get id(\otherid)
        
        \pgfooget{child}{\selfchild}
        \ifnum\selfchild=0
            \expanded{%
              \noexpand\pgfoonew\expandafter\noexpand\csname #2\endcsname=new Value(\newdata,{\selfid,\otherid},+,0)%
            }%
            \csname #2\endcsname.get id(\childid)
            \pgfooeset{child}{\childid}
        \else
            \pgfooobj{\selfchild}.setdata(\newdata)
            \pgfooobj{\selfchild}.get handle(\childobj)
            \expandafter\let\csname #2\endcsname = \childobj
        \fi
    }

    \method relu(#1) { 
        \pgfooget{data}{\selfdata}

        \pgfoothis.get id(\selfid)

        \FPifpos\selfdata
            \edef\newdata{\selfdata}
        \else
            \edef\newdata{0}
        \fi
        
        \pgfooget{next}{\nextIDs}

        \ifx\nextIDs\empty
            \expanded{%
              \noexpand\pgfoonew\expandafter\noexpand\csname #1\endcsname=new Value(\newdata,{\selfid},relu,0)%
            }%
            \csname #1\endcsname.get id(\nextID)
            \pgfooeset{next}{\nextID}
        \else
            \splicelist{\nextIDs}
            \pgfooobj{\myhead}.setdata(\newdata)
            \pgfooobj{\myhead}.get handle(\nextobj)
            \expandafter\let\csname #1\endcsname = \nextobj
        \fi
    }

    \method localbackwards() {
    
        \pgfooget{grad}{\currentgrad}

        \pgfooget{op}{\currentop}

        \pgfoothis.get id(\localtempid)
        
        \pgfooget{prev}{\localparentids}

        \debug{Value: \localtempid\space - Inside LocalBackwards: has grad \currentgrad\space and prev \localparentids}{HIGH}

        \ifnum\pdfstrcmp{\currentop}{+}=0
            
            \expandafter\splitlist\localparentids\relax 
            \pgfooobj{\firstid}.getgrad(\firstgrad)
            \FPeval\newfirstgrad{(\firstgrad) + (\currentgrad)}
            \pgfooobj{\firstid}.setgrad(\newfirstgrad)
            \pgfooobj{\firstid}.increasegradcounter()

            \debug{Value: \localtempid\space - Inside LocalBackwards ADD: Updating \firstid's grad using \firstgrad\space + \currentgrad}{HIGH}
            
            \pgfooobj{\secondid}.getgrad(\secondgrad)
            \FPeval\newsecondgrad{(\secondgrad) + (\currentgrad)}
            \pgfooobj{\secondid}.setgrad(\newsecondgrad)
            \pgfooobj{\secondid}.increasegradcounter()

            \debug{Value: \localtempid\space - Inside LocalBackwards ADD: Updating \secondid's grad using \secondgrad\space + \currentgrad}{HIGH}
        \else
            \ifnum\pdfstrcmp{\currentop}{*}=0
                
                \expandafter\splitlist\localparentids\relax 
    
                \pgfooobj{\firstid}.getdata(\firstdata)
                \pgfooobj{\secondid}.getdata(\seconddata)
                
                \FPeval\newfirstgrad{(\currentgrad) * (\seconddata)} 
                \pgfooobj{\firstid}.getgrad(\oldfirstgrad)
                \FPeval\updatedfirstgrad{(\oldfirstgrad) + (\newfirstgrad)}
                \pgfooobj{\firstid}.setgrad(\updatedfirstgrad)
                \pgfooobj{\firstid}.increasegradcounter()
                
                \debug{Value: \localtempid\space - Inside LocalBackwards MUL: Updating \firstid's grad using \oldfirstgrad\space + (\currentgrad) * (\seconddata)}{HIGH}
                \pgfooobj{\firstid}.debugmsg(HIGH)
                
                \FPeval\newsecondgrad{(\currentgrad) * (\firstdata)}
                \pgfooobj{\secondid}.getgrad(\oldsecondgrad)
                \FPeval\updatedsecondgrad{(\oldsecondgrad) + (\newsecondgrad)}
                \pgfooobj{\secondid}.setgrad(\updatedsecondgrad)
                \pgfooobj{\secondid}.increasegradcounter()
    
                \debug{Value: \localtempid\space - Inside LocalBackwards MUL: Updating \secondid's grad using \oldsecondgrad\space + (\currentgrad) * (\firstdata)}{HIGH}
                \pgfooobj{\secondid}.debugmsg(HIGH)

            \else
                \ifnum\pdfstrcmp{\currentop}{relu}=0 
                    \pgfooget{data}{\selfdata}
                    \pgfooget{grad}{\currentgrad}
        
                    \pgfooget{prev}{\parentid}
        
                    \debug{Value: \localtempid\space - Inside LocalBackwards ReLU}{HIGH}

                    \def\minvalue{0.0}
                    \ifnum\fpeval{\selfdata > \minvalue} = 1
                        \pgfooobj{\parentid}.getgrad(\parentgrad)
                        \edef\expandedvar{\parentgrad}
                        \let\parentgrad\expandedvar
                        
                        \FPeval\newparentgrad{(\parentgrad) + (\currentgrad)}
                        \edef\expandedvar{\newparentgrad}
                        \let\newparentgrad\expandedvar
                        \pgfooobj{\parentid}.setgrad(\newparentgrad)
                        \pgfooobj{\parentid}.increasegradcounter()
                    \else
                        \debug{Value: \localtempid\space - Inside LocalBackwards ReLU: Data negative so not backpropagating}{HIGH}
                        \pgfooobj{\parentid}.increasegradcounter()
                    \fi
                \else
                    \ifnum\pdfstrcmp{\currentop}{''}=0
                        \debug{Value: \localtempid\space - Inside LocalBackwards: This is a leaf node, nothing to pass back}{HIGH}
                    \else
                        \debug{Value: \localtempid\space - Inside LocalBackwards: Operation: \currentop\space  Not Yet Implemented}{LOW}
                        \pgfoothis.debugmsg(LOW)
                    \fi
                \fi
            \fi
        \fi

        \pgfoothis.setdonebackwards(1.0)

    }

    \method backward() {
        
        \pgfooset{grad}{1.0}
        \pgfoothis.get id(\currentid)

        \pgfoothis.getcomputegraph(\tempcg)
        \listlength{\cglen}{\tempcg}

        \debug{Value: \currentid\space - Inside Backward, Compute graph length is \cglen}{MED}

        \ifthenelse{\cglen=0}{
            \debug{Value: \currentid\space - Inside Backward, Static Compute Graph Not Computed}{MED}
            \initQueue{queue}
            \enqueue{queue}{\currentid}
    
            \loop
                \dequeue{queue}{\tempNode} 
                
                \debug{Value: \currentid\space - Inside Backward, Removed: \tempNode}{HIGH}
    
                \pgfooobj{\tempNode}.debugmsg(MED)
                \pgfooobj{\tempNode}.getdata(\tempData)
                \pgfooobj{\tempNode}.getop(\tempOp)
                \pgfooobj{\tempNode}.getprev(\tempParents)
                \pgfooobj{\tempNode}.getgrad(\tempGrad)
                \pgfooobj{\tempNode}.getnext(\tempNextIDs)
                \pgfooobj{\tempNode}.getgradcounter(\tempGradCounter)
                \pgfooobj{\tempNode}.getdonebackwards(\tempDoneBackwards)
        
                \listlength{\numNextIDs}{\tempNextIDs}

                \ifnum\fpeval{\tempGradCounter - \numNextIDs} = 0
                    \ifnum\fpeval{\tempDoneBackwards - 0.0} = 0
                        \pgfoothis.appendtocomputegraph(\tempNode)
                        \pgfooobj{\tempNode}.localbackwards()
                        \ifthenelse{\pdfstrcmp{\tempParents}{}=0}{
                            \debug{Value: \currentid\space - Inside Backward, No parents to add to the queue}{HIGH}
                        }{
                            \ifthenelse{\pdfstrcmp{\tempOp}{relu}=0}{
                                \debug{Value: \currentid\space - Inside Backward, Adding single parent \tempParents\space onto the queue}{HIGH}
                                
                                \enqueue{queue}{\tempParents}
                            }{
                                \debug{Value: \currentid\space - Inside Backward, Adding parents onto the queue}{HIGH}
                                
                                \expandafter\splitlist\tempParents\relax 
                                
                                \enqueue{queue}{\firstid}
                                \enqueue{queue}{\secondid}
                            }
                        }
                    \else
                        \debug{Value: \currentid\space - Inside Backward, Already called local Backwards on this Node}{HIGH}
                    \fi
                \else
                    \debug{Value: \currentid\space - Inside Backward, Adding \tempNode\space back onto queue as not all gradients received yet}{HIGH}
                    \debug{Value: \currentid\space - Inside Backward, Node \tempNode\space has GradCounter \tempGradCounter\space and \numNextIDs\space parents}{HIGH}
                    \enqueue{queue}{\tempNode}
                \fi
    
                \ifnum\csname queue@head\endcsname<\csname queue@tail\endcsname\repeat
        } {
            \debug{Value: \currentid\space - Inside Backward, Static Compute Graph Already Computed}{MED}
            
            \splicelist{\tempcg}

            \pgfplotsforeachungrouped \iinbackward in {1,...,\cglen} {

                \debug{Value: \currentid\space - Inside Backward, Calling localbackwads on Value \myhead}{HIGH}
                
                \pgfooobj{\myhead}.localbackwards()

                \splicelist{\mytail}
            }
        }
    }

    \method zerononparams() {
        \pgfoothis.get id(\currentid)

        \pgfoothis.getcomputegraph(\tempcg)
        \listlength{\cglen}{\tempcg}

        \debug{Value: \currentid\space - Inside ZeroNonParams, Compute graph length is \cglen}{MED}

        \ifthenelse{\cglen=0}{
            LOG: Must run compile() or backward() once before zerononparams()
        }{
            \debug{Value: \currentid\space - Inside ZeroNonParams, Static Compute Graph Already Computed}{MED}
            
            \splicelist{\tempcg}

            \pgfplotsforeachungrouped \iinzerononparams in {1,...,\cglen} {

                \pgfooobj{\myhead}.getisparam(\tempIsparam)

                \ifthenelse{\tempIsparam=0}{
                    \pgfooobj{\myhead}.setgrad(0.0)
                    \pgfooobj{\myhead}.setgradcounter(0.0)
                    \pgfooobj{\myhead}.setdonebackwards(0.0)
                }{}              

                \splicelist{\mytail}
            }
        }
    }

    \method zero() {
        \pgfoothis.get id(\currentid)

        \pgfoothis.getcomputegraph(\tempcg)
        \listlength{\cglen}{\tempcg}

        \debug{Value: \currentid\space - Inside Zero, Compute graph length is \cglen}{MED}

        \ifthenelse{\cglen=0}{
            LOG: Must run compile() or backward() once before zero()
        }{
            \debug{Value: \currentid\space - Inside Zero, Static Compute Graph Already Computed}{MED}
            
            \splicelist{\tempcg}

            \pgfplotsforeachungrouped \iinzero in {1,...,\cglen} {

                \debug{Value: \currentid\space - Inside Zero, Resetting Value \myhead}{HIGH}
                
                \pgfooobj{\myhead}.setgrad(0.0)
                \pgfooobj{\myhead}.setgradcounter(0.0)
                \pgfooobj{\myhead}.setdonebackwards(0.0)

                \splicelist{\mytail}
            }
        }
    }

    \method step(#1) {
        \pgfoothis.get id(\currentid)

        \pgfoothis.getcomputegraph(\tempcg)
        \listlength{\cglen}{\tempcg}

        \debug{Value: \currentid\space - Inside step, Compute graph length is \cglen}{MED}

        \ifthenelse{\cglen=0}{
            LOG: Must run compile() or backward() once before step()
        }{
            \debug{Value: \currentid\space - Inside step, Static Compute Graph Already Computed}{MED}
            
            \splicelist{\tempcg}

            \pgfplotsforeachungrouped \iinstep in {1,...,\cglen} {
                
                \pgfooobj{\myhead}.getdata(\tempData)
                \pgfooobj{\myhead}.getgrad(\tempGrad)
                \pgfooobj{\myhead}.getisparam(\tempIsparam)

                \ifthenelse{\tempIsparam=1}{
                    \debug{Value: \currentid\space - Inside step, Updating Value \myhead\space with data \tempData\space grad \tempGrad\space lr #1}{HIGH}
                    \FPeval\newval{(\tempData) - (#1) * (\tempGrad)}
                    \pgfooobj{\myhead}.setdata(\newval)
                    \debug{Value: \currentid\space - Inside step, New data is \newval}{HIGH}
                }{}

                \splicelist{\mytail}
            }
        }
    }
}

\pgfooclass{Module}{

    \method Module() {
    
    }
}

\pgfooclass(Module){Neuron}{
    \attribute numIn;
    \attribute w; 
    \attribute b; 
    \attribute nonlin;
    \attribute dummy;

    \method Neuron(#1,#2) {
        \pgfooeset{numIn}{#1}

        \pgfoonew \dummy=new Value(0,{},'',0) 
        \dummy.get id(\dummyID)
        \pgfooeset{dummy}{\dummyID}

        \def\weightIDs{}
        
        \pgfplotsforeachungrouped \i in {1,...,#1} {
            \boundrandfloat{\tempRand}{-1.0}{1.0}
            
            \pgfoonew \tempValue=new Value(\tempRand,{},'',1)
            \tempValue.get id(\tempID)

            \ifx\weightIDs\empty
                \edef\weightIDs{\tempID}
            \else
                \edef\weightIDs{\weightIDs,\tempID}
            \fi
        }

        \pgfooeset{w}{\weightIDs}

        \pgfoonew \b=new Value(0,{},'',1) 

        \b.get id(\tempID)
        \pgfooeset{b}{\tempID}
        
        \pgfooeset{nonlin}{#2}
    }

    \method show() {
        \pgfooget{numIn}{\tempnumIn}
        \pgfooget{b}\tempb
        \pgfooget{nonlin}{\tempnonlin}
        \pgfoothis.get id(\tempid)
        \pgfooget{w}{\tempw}
        Neuron(self: \tempid, NumIn: \tempnumIn, Bias: \tempb, Activation: \tempnonlin, Weights IDs: \tempw)\par
    }

    \method debugmsg(#1) {
        \pgfooget{numIn}{\tempnumIn}
        \pgfooget{b}\tempb
        \pgfooget{nonlin}{\tempnonlin}
        \pgfoothis.get id(\tempid)
        \pgfooget{w}{\tempw}
        \debug{Neuron(self: \tempid, NumIn: \tempnumIn, Bias: \tempb, Activation: \tempnonlin, Weights IDs: \tempw)}{#1}
    }

    \method forward(#1,#2) {
        \pgfooget{w}{\weightIDs}
        \pgfooget{numIn}{\numIn}
        \pgfooget{dummy}{\dummyID}
        \pgfooobj{\dummyID}.get handle(\runningsum)
        \pgfooget{nonlin}{\nonlin}

        \zip*{/}#1\weightIDs\prodlist
        \splicelist{\prodlist}

        \pgfplotsforeachungrouped \i in {1,...,\numIn} {

            \edef\neuronForwardNextTail{\mytail}

            \expandafter\splitstringslash\myhead\relax

            \debug{Neuron: - Inside Forward, MyHead: \myhead\space , MyTail: \mytail}{MED}
            \debug{Neuron: - Inside Forward, Multiplying Input ID: \inputID\space by Weight ID: \weightID}{MED}

            \pgfooobj{\weightID}.multiply(\pgfooobj{\inputID},prod)
            \prod.debugmsg(MED)
            \prod.add(\runningsum,runningsum)

            \splicelist{\neuronForwardNextTail}
        }

        \debug{Neuron: - Inside Forward, Output of weight matrix:}{MED}
        \runningsum.debugmsg(MED)

        \pgfooget{b}{\biasID}

        \ifthenelse{\equal{\nonlin}{relu}}{
            \debug{Neuron: - Inside Forward, Adding Bias \biasID\space with value: }{HIGH}
            \pgfooobj{\biasID}.debugmsg(HIGH)
            \runningsum.add(\pgfooobj{\biasID},output)
            \debug{Neuron: - Inside Forward, To give: }{HIGH}
            \output.debugmsg(HIGH)
            \debug{Neuron: - Inside Forward, Applying ReLU}{HIGH}
            \output.relu(#2)
        }{
            \runningsum.add(\pgfooobj{\biasID},#2)
            \debug{Neuron: - Inside Forward, Adding Bias without ReLU}{HIGH}
        }
    }

    \method getparameters(#1) {
        \pgfooget{w}{\weightIDs}
        \pgfooget{b}{\biasID}
        \edef\neuronparamlist{\weightIDs,\biasID}
        \expandafter\let\csname #1\endcsname = \neuronparamlist
    }
}

\pgfooclass(Module){Layer}{
    \attribute numIn;
    \attribute numOut;
    \attribute nonlin;
    \attribute neurons; 

    \method Layer(#1,#2,#3) {
        \pgfooeset{numIn}{#1}
        \pgfooeset{numOut}{#2}
        \pgfooeset{nonlin}{#3}

        \def\neuronIDs{}
        
        \pgfplotsforeachungrouped \i in {1,...,#2} {
            \pgfoonew \tempNeuron=new Neuron(#1,#3)
            \tempNeuron.get id(\tempID)
            \tempNeuron.debugmsg(MED)
            
            \ifx\neuronIDs\empty
                \edef\neuronIDs{\tempID}
            \else
                \edef\neuronIDs{\neuronIDs,\tempID}
            \fi
        }
        \pgfooeset{neurons}{\neuronIDs}
    }

    \method forward(#1,#2) {
        \pgfooget{numOut}{\numOut}
        \pgfooget{neurons}{\neuronIDs}
        \splicelist{\neuronIDs}

        \def\outputIDs{}
    
        \pgfplotsforeachungrouped \k in {1,...,\numOut} {
            \debug{Layer: - Inside Forward, \k\space of \numOut}{HIGH}
            \edef\currentHead{\myhead}
            \edef\currentTail{\mytail}
    
            \pgfooobj{\currentHead}.forward(#1,tempNeuronOutput)
    
            \tempNeuronOutput.get id(\tempOutputID)

            \debug{Layer: - Inside Forward, Output of Neuron: }{HIGH}
            \tempNeuronOutput.debugmsg(HIGH)
    
            \ifx\outputIDs\empty
                \edef\outputIDs{\tempOutputID}
            \else
                \edef\outputIDs{\outputIDs,\tempOutputID}
            \fi

            \splicelist{\currentTail}
        }

        \expandafter\let\csname #2\endcsname = \outputIDs
    }

    \method show() {
        \pgfooget{numIn}{\tempnumIn}
        \pgfooget{numOut}{\tempnumOut}
        \pgfoothis.get id(\tempid)
        \pgfooget{neurons}{\tempneurons}
        Layer(self: \tempid, NumIn: \tempnumIn, NumOut: \tempnumOut, Neuron IDs: \tempneurons)\par
    }

    \method debugmsg(#1) {
        \pgfooget{numIn}{\tempnumIn}
        \pgfooget{numOut}{\tempnumOut}
        \pgfoothis.get id(\tempid)
        \pgfooget{neurons}{\tempneurons}
        \debug{Layer(self: \tempid, NumIn: \tempnumIn, NumOut: \tempnumOut, Neuron IDs: \tempneurons)}{#1}
    }

    \method getparameters(#1) {
        \pgfooget{neurons}{\neuronIDs}
        \pgfooget{numOut}{\numOut}
        \def\allparams{}
        \splicelist{\neuronIDs}
        \pgfplotsforeachungrouped \i in {1,...,\numOut} {
            \edef\currentHead{\myhead}
            \edef\currentTail{\mytail}
            \pgfooobj{\currentHead}.getparameters(paramlist)
            \ifx\allparams\empty
                \edef\allparams{\paramlist}
            \else
                \edef\allparams{\allparams,\paramlist}
            \fi
            \splicelist{\currentTail}
        }
        \expandafter\let\csname #1\endcsname = \allparams
    }
}

\pgfooclass(Module){MLP}{
    \attribute numIn;
    \attribute layerOutputSizes; 
    \attribute numLayers;
    \attribute layers; 

    \method MLP(#1,#2) {
        \pgfooeset{numIn}{#1}
        \pgfooeset{layerOutputSizes}{#2}

        \listlength{\len}{#2}
        \pgfooeset{numLayers}{\len}
        
        \splicelist{#2}

        \def\layerIDs{}
        
        \pgfplotsforeachungrouped \i in {1,...,\len} {
            \edef\currentHead{\myhead}
            \edef\currentTail{\mytail}            

            \ifthenelse{\i=1}{
                \debug{MLP: - Inside Init, Creating First Layer with input: #1 and output: \currentHead}{MED}
                \pgfoonew \tempLayer=new Layer(#1,\currentHead,relu)
            } {
                \splicelist{\mytail}
                \edef\nextHead{\myhead}
                    
                \ifthenelse{\i=\len}{
                    \debug{MLP: - Inside Init, Creating last Layer with input: \currentHead\space and output: \nextHead}{MED}
                    \pgfoonew \tempLayer=new Layer(\currentHead,\nextHead,none)
                } {
                    \debug{MLP: - Inside Init, Creating hidden Layer with input: \currentHead\space and output: \nextHead}{MED}
                    \pgfoonew \tempLayer=new Layer(\currentHead,\nextHead,relu)
                }

                \splicelist{\currentTail}
            }
            
            \tempLayer.get id(\tempID)
            \tempLayer.debugmsg(MED)
            
            \ifx\layerIDs\empty
                \edef\layerIDs{\tempID}
            \else
                \edef\layerIDs{\layerIDs,\tempID}
            \fi
        }
        \pgfooeset{layers}{\layerIDs}
    }

    \method forward(#1,#2) {
        
        \pgfooget{layers}{\layerIDs}
        \pgfooget{numLayers}{\numLayers}
        \splicelist{\layerIDs}

        \pgfplotsforeachungrouped \j in {1,...,\numLayers} {

            \edef\currentLayerHead{\myhead}
            \edef\currentLayerTail{\mytail}
            
            \debug{MLP: - Inside Forward, Current Layer ID: \currentLayerHead}{HIGH}
            \debug{MLP: - Inside Forward, Remaining Layer IDs: \currentLayerTail}{HIGH}
    
            \ifthenelse{\j=1}{
                \pgfooobj{\currentLayerHead}.forward(#1,tempLayerOutput)
            } {
                \pgfooobj{\currentLayerHead}.forward(\tempLayerOutput,tempLayerOutput)
            }

            \debug{MLP: - Inside Forward, Layer Output IDs: \tempLayerOutput}{HIGH}
            \ifthenelse{\j=\numLayers}{
                
            } {}

            \splicelist{\currentLayerTail}

        }

        \pgfooobj{\tempLayerOutput}.get handle(\layeroutputobj)
        \expandafter\let\csname #2\endcsname = \layeroutputobj
    }

    \method show() {
        \pgfoothis.get id(\tempid)
        \pgfooget{numIn}{\tempnumIn}
        \pgfooget{numLayers}{\tempnumLayers}
        \pgfooget{layerOutputSizes}{\templayerOutputSizes}
        \pgfooget{layers}{\templayerIDs}
        MLP(self: \tempid, NumIn: \tempnumIn, NumLayers: \tempnumLayers, LayerOutputSizes: \templayerOutputSizes, Layer IDs: \templayerIDs)\par
    }

    \method debugmsg(#1) {
        \pgfoothis.get id(\tempid)
        \pgfooget{numIn}{\tempnumIn}
        \pgfooget{numLayers}{\tempnumLayers}
        \pgfooget{layerOutputSizes}{\templayerOutputSizes}
        \pgfooget{layers}{\templayerIDs}
        \debug{MLP(self: \tempid, NumIn: \tempnumIn, NumLayers: \tempnumLayers, LayerOutputSizes: \templayerOutputSizes, Layer IDs: \templayerIDs)}{#1}
    }

    \method getparameters(#1) {
        \pgfooget{layers}{\layerIDs}
        \listlength{\numLayers}{\layerIDs}
        \def\allMLPparams{}
        \splicelist{\layerIDs}
        \pgfplotsforeachungrouped \i in {1,...,\numLayers} {
            \edef\layerIDHead{\myhead}
            \edef\layerIDTail{\mytail}
            \pgfooobj{\layerIDHead}.getparameters(MLPparamlist)
            \ifx\allMLPparams\empty
                \edef\allMLPparams{\MLPparamlist}
            \else
                \edef\allMLPparams{\allMLPparams,\MLPparamlist}
            \fi
            \splicelist{\layerIDTail}
        }
        \expandafter\let\csname #1\endcsname = \allMLPparams
    }
}

\newcommand{\spiral}[2]{
    
    \def\spiraldataset{}
    
    \pgfmathsetmacro{\a}{0.5}         
    \pgfmathsetmacro{\b}{0.5}       
    \pgfmathsetmacro{\maxangle}{2}  
    \pgfmathsetmacro{\noise}{0.5}   
        
    \pgfplotsforeachungrouped \i in {1,...,#2} {
        \pgfmathsetmacro{\theta}{random() * \maxangle * pi}
        \pgfmathsetmacro{\r}{\a + \b * \theta}
        \pgfmathsetmacro{\x}{\r * cos(deg(\theta)) + (random() - 0.5) * \noise}
        \pgfmathsetmacro{\y}{\r * sin(deg(\theta)) + (random() - 0.5) * \noise}

        \ifx\spiraldataset\empty
            \edef\spiraldataset{(\x,\y,-1)}
        \else
            \edef\spiraldataset{\spiraldataset,(\x,\y,-1)}
        \fi

        \pgfmathsetmacro{\theta}{random() * \maxangle * pi}
        \pgfmathsetmacro{\r}{\a + \b * \theta}
        \pgfmathsetmacro{\x}{\r * cos(deg(\theta + pi)) + (random() - 0.5) * \noise}
        \pgfmathsetmacro{\y}{\r * sin(deg(\theta + pi)) + (random() - 0.5) * \noise}

        \edef\spiraldataset{\spiraldataset,(\x,\y,1)}
        
    }

    \expandafter\let\csname #1\endcsname\spiraldataset

}

\newcommand{\splittuple}[1]{%
  \expandafter\splittupleaux#1%
}
\def\splittupleaux(#1,#2,#3){%
  \def\inputx{#1}%
  \def\inputy{#2}%
  \def\target{#3}%
}

\def\firsttuple(#1,#2,#3),#4\relax{
  \def\inputx{#1} 
  \def\inputy{#2} 
  \def\target{#3} 
  \def\rest{#4}   
}

\def\lasttuple(#1,#2,#3)\relax{%
  \def\inputx{#1}%
  \def\inputy{#2}%
  \def\target{#3}%
  \def\rest{}%
}

\newcommand{\displayspiral}[1]{
    \begin{tikzpicture}[scale=1, every node/.style={scale=1.0}]
    \foreach \datasetitem in #1 {
        \splittuple{\datasetitem}
        \ifnum\target=-1
            \fill[red] (\inputx, \inputy) circle (2pt);
        \else
            \fill[blue] (\inputx, \inputy) circle (2pt);
        \fi
    }

    \draw[->, thick] (-5, 0) -- (5, 0) node[right] {$x$};
    \draw[->, thick] (0, -5) -- (0, 5) node[above] {$y$};

     \foreach \x in {-5,-4,...,-1,0,1,...,4,5} {
       \draw (\x,0.1) -- (\x,-0.1); 
     }
     \foreach \y in {-5,-4,...,-1,0,1,...,4,5} {
       \draw (0.1,\y) -- (-0.1,\y);
     }

     \foreach \x/\i in {0/0, 1/1, 2/2} {
       \node[below] at (\x, -0.2) {$\i$};
     }
     \foreach \y/\i in {0/0, 1/1, 2/2} {
       \node[left] at (-0.2, \y) {$\i$};
     }

    \end{tikzpicture}
}

\newcommand{\findminmax}[5]{
    
    \def\tempxmin{10000}\def\tempxmax{-10000}
    \def\tempymin{10000}\def\tempymax{-10000}
    
    \edef\datasetcopy{\csname #1\endcsname}
    \loop
        \ifx\datasetcopy\empty
        \else
            \expandafter\firsttuple\datasetcopy\relax
            \edef\datasetcopy{\rest}
            
            \pgfmathsetmacro{\temp}{\x < \tempxmin ? \x : \tempxmin}
            \xdef\tempxmin{\temp}
            \pgfmathsetmacro{\temp}{\x > \tempxmax ? \x : \tempxmax}
            \xdef\tempxmax{\temp}
            
            \pgfmathsetmacro{\temp}{\y < \tempymin ? \y : \tempymin}
            \xdef\tempymin{\temp}
            \pgfmathsetmacro{\temp}{\y > \tempymax ? \y : \tempymax}
            \xdef\tempymax{\temp}
    \repeat
    
    \pgfmathsetmacro{\temp}{\tempxmin - 1}
    \expandafter\xdef\csname #2\endcsname{\temp}
    \pgfmathsetmacro{\temp}{\tempxmax + 1}
    \expandafter\xdef\csname #3\endcsname{\temp}
    \pgfmathsetmacro{\temp}{\tempymin - 1}
    \expandafter\xdef\csname #4\endcsname{\temp}
    \pgfmathsetmacro{\temp}{\tempymax + 1}
    \expandafter\xdef\csname #5\endcsname{\temp}
}

\newcommand{\createmesh}[6]{
    
    \def\meshgrid{}
    
    \pgfmathsetmacro{\xsteps}{int((#4 - #3) / #2 + 1)}
    \pgfmathsetmacro{\ysteps}{int((#6 - #5) / #2 + 1)}
    
    \pgfplotsforeachungrouped \i in {0,...,\xsteps} {
        \pgfplotsforeachungrouped \j in {0,...,\ysteps} {
            \pgfmathsetmacro{\xcurr}{#3 + \i * #2}
            \pgfmathsetmacro{\ycurr}{#5 + \j * #2}
            
            \ifx\meshgrid\empty
                \edef\meshgrid{(\xcurr,\ycurr)}
            \else
                \edef\meshgrid{\meshgrid,(\xcurr,\ycurr)}
            \fi
        }
    }
    
    \expandafter\let\csname #1\endcsname\meshgrid
} 
\newwrite\outfile  

\setboolean{debug}{false} 
\newcommand{\debuglevel}{LOW} 

\newboolean{resume}

\newcommand{\resumetraining}[1]{
	\newread\infile  
	\def\mlpweights{}    
	\def\startepoch{}

	\openin\infile=#1  
	\read\infile to \mlpweights    
	\read\infile to \startepoch    
	\read\infile to \losses    
	\closein\infile            
	\edef\startepoch{\the\numexpr\startepoch+1\relax}
	\message{Loaded weights: \mlpweights^^J}
	\message{Start epoch: \startepoch^^J}
	\message{Losses: \losses^^J}
	
	\message{Setting loaded weights^^J}
	\zip*{/}\mlpparams\mlpweights\mlpweightsandids
	\splicelist{\mlpweightsandids}
	\pgfplotsforeachungrouped \i in {1,...,\len} {
		\edef\loadNextHead{\myhead}
            	\edef\loadNextTail{\mytail}
		\expandafter\splitstringslash\myhead\relax
		\pgfooobj{\inputID}.setdata(\weightID)
		\splicelist{\loadNextTail}
	}

}

\newcommand{\savecheckpoint}[3]{
	\def\mlpweights{}
	\foreach \i in #1 {
	    \pgfooobj{\i}.getdata(\tempData)
	    \ifx\mlpweights\empty
		\xdef\mlpweights{\tempData}
	    \else
	        \xdef\mlpweights{\mlpweights,\tempData}
	    \fi        
	}

	\message{Saving checkpoint^^J}

	\message{Weights: \mlpweights^^J}

	\immediate\openout\outfile=checkpoint.txt  
	\immediate\write\outfile{\mlpweights}      
	\immediate\write\outfile{#2}      
	\immediate\write\outfile{#3}      
	\immediate\closeout\outfile            

}

\makeatletter
\def\logosmaller#1{%
  \hbox{\sbox\z@ T%
    \vbox to\ht\z@{\hbox{\check@mathfonts
      \fontsize\sf@size\z@
      \math@fontsfalse\selectfont #1}\vss}}}
      
\newcommand\NeuRaLaTeX{%
  N\kern-.1em%
  \raise-0.5ex\hbox{E}\kern-.1em%
  \logosmaller{U}\kern-.05em%
  R\kern-.1em%
  \raise-0.5ex\hbox{A}\kern-.1em%
  \LaTeX
  }

\begin{document}

\title{\NeuRaLaTeX: a machine learning \\ library written in pure \LaTeX}

\author{James A.~D.~Gardner, Will Rowan and William A.~P.~Smith\\
Department of Computer Science\\
University of York, UK
}

\maketitle

\def\numberofepochs{35}
\def\trainitemsperclass{50}
\def\evalitemsperclass{50}
\def\batchsize{100}
\setboolean{resume}{true}
\def\checkpointfile{checkpoint35epochs.txt}


\begin{abstract}
In this paper, we introduce \NeuRaLaTeX, which we believe to be the first deep learning library written entirely in \LaTeX. As part of your \LaTeX~document you can specify the architecture of a neural network and its loss functions, define how to generate or load training data, and specify training hyperparameters and experiments. When the document is compiled, the \LaTeX~compiler will generate or load training data, train the network, run experiments, and generate figures. This paper generates a random 100 point spiral dataset, trains a two layer MLP on it, evaluates on a different random spiral dataset, produces plots and tables of results. The paper took 48 hours to compile and the entire source code for \NeuRaLaTeX~is contained within the source code of the paper.

We propose two new metrics: the Written In Latex (WIL) metric measures the proportion of a machine learning library that is written in pure \LaTeX, while the Source Code Of Method in Source Code of Paper (SCOMISCOP) metric measures the proportion of a paper's implementation that is contained within the paper source. We are state-of-the-art for both metrics, outperforming the ResNet and Transformer papers, as well as the PyTorch and Tensorflow libraries. Source code, documentation, videos, crypto scams and an invitation to invest in the commercialisation of \NeuRaLaTeX~are available at \href{https://www.neuralatex.com}{neuralatex.com}.
\end{abstract}

\section{Introduction}

While often used as a document preparation markup language, in fact \LaTeX~is itself a Turing-complete programming language. \LaTeX~natively provides variables, loops and conditionals via \TeX~primitives, while other features such as object-orientation are supported via appropriate packages \cite{tantau2024tikz}. Yet, if you consult the curated list of awesome machine learning frameworks, libraries and software (by language) \cite{awesome}, \LaTeX~is not even included as a category of programming language. This motivated us to write a deep learning library entirely in \LaTeX.

\subsection{Wait, what?}

When we talk about `compiling' a \LaTeX~document, really what we mean is that the program written in your \LaTeX~document is executed. The \LaTeX~engine parses and processes the document, resolving macros, executing conditionals, and formatting content before outputting the final result. \NeuRaLaTeX~is a collection of \texttt{.tex} \LaTeX~source files that you can include in your document via \texttt{{\textbackslash}input\{...\}}. These provide a series of commands that enable you to define a neural network, load or generate training and evaluation data, train the network via backprop and run inference on the trained network. All of this happens when your document is `compiled'. The results shown in Section \ref{sec:evaluation} are all dynamically generated at compile time using a neural network that was itself trained while the \LaTeX~document compiled.

\subsection{Why \LaTeX?}

\subsubsection{An ideal programming language}

Do you sometimes find it hard to decide whether to expand, not expand or expand after? Well, with \LaTeX~you can do all of those and more (see Figure \ref{fig:expandnoexpand}). Who needs variables when you have macros? Who needs arrays when you can create comma, separated, strings? Who wants a simple \texttt{for} loop when you can have \texttt{pgfplotsforeachungrouped}?

\begin{figure}[!t]
    \centering
    \begin{minipage}{\columnwidth}
        \lstset{
            language={[LaTeX]TeX}, 
            basicstyle=\ttfamily\small, 
            keywordstyle=\color{blue}, 
            commentstyle=\color{gray}, 
            frame=single, 
            breaklines=true 
        }
\begin{lstlisting}
\expanded{
  \noexpand\pgfoonew\expandafter\noexpand
  \csname #2\endcsname=
  new Value(\newdata,{\selfid,\otherid},*,0)
}
\end{lstlisting} 
    \end{minipage}
    \caption{A pure \LaTeX~implementation has the benefit of high code readability. For example, returning the result of a product between two value objects only uses the word `expand' four times.}
    \label{fig:expandnoexpand}
\end{figure}

\subsubsection{The self-contained paper}

Machine learning is facing a reproducibility crisis. Too often \emph{``code will be made available upon paper acceptance''} becomes a GitHub repository that is empty but for a README.md containing words that strike fear into the hearts of PhD students hoping for an easy comparative evaluation: \emph{``Code coming soon''}.

On the other hand, any machine learning researcher worth their salt uploads all of their papers to arXiv. The requirement by arXiv for all papers to submit the full \LaTeX~source necessary to compile the paper presents an opportunity. In this paper, we exploit the universality of access to paper source files as a solution to the reproducibility problem. We call this the \emph{self-contained paper}. A self-contained paper must contain within its \LaTeX~source all training data, implementation of the method, and experiments in a form that can be run by the \LaTeX~compiler itself. The training of any models therefore takes place as part of the compilation of the paper. Since arXiv makes the paper source files available, having access to the paper is equivalent to having access to the code. No more ``code coming soon''!

\subsubsection{Additional benefits}

First, overleaf becomes more than just a cloud-based \LaTeX~editor. It is now also your (free) cloud compute service. Second, since arXiv limits the size of any submission to 50MB, researchers are forced to work on very small datasets and models. This helps reduce the unfair advantage industry has in accessing large GPU resources. Third, \LaTeX~is a programming language, so why have to context switch between your paper source and Python IDE - just do everything in one place (\LaTeX)! Finally, do you always forget git commands? From now on, your code link can simply point to the paper source files on arXiv. No need for a GitHub repository (who needs version control anyway?).

\subsection{Why?}

More seriously, implementing a neural network library in such an awkward `programming language' has been an incredible learning experience. You might think you understand backprop, but actually implementing it from scratch in a programming language that lacks most of the basic features you rely on in any other language is a seriously fun and intellectually challenging exercise. It's also worth emphasising that none of the authors are particularly knowledgable about \LaTeX. It's quite possible that we made life much harder for ourselves than necessary. We were never fully confident about the scope of macros, so we had to use defensive naming conventions in case they were global. Another interesting challenge is that LLMs like ChatGPT are pretty terrible at programming in \LaTeX~so help is limited. If you ask them how to do something complicated in \LaTeX, they tend to politely suggest you use python instead and provide python code. Or if they do provide \LaTeX~ code it often doesn't work.

\subsection{Related work}

Important previous work has also considered implementation of different types of program in neglected languages. For example, Wildenhain \cite{wildenhain2017on} showed that MS PowerPoint is Turing-complete, providing a cross-platform, intuitive, GUI-based programming language in which any conceivable program could be implemented. Closely related to our concept of unifying both the implementation of a method with the source code of its write-up, Murphy \cite{murphy2017zm} showed how a single file could simultaneously be both a valid executable file and also a plain text file containing the paper itself. Like us, Wildenhain \cite{wildenhain2018wordtex} also understand the superiority of \LaTeX, but rather than make \LaTeX~more powerful, they dumb it down to a WYSIPCTWOTCG (What You See Is Pretty Close To What Other Tools Can Get) editor, WordTeX. The most closely related previous work is ExcelNet \cite{fouhey2016excel} that implemented neural networks in Microsoft Excel. However, they did not implement backpropagation and only supported pretrained (or user edited) network weights. Inspiration for our catchy name came from ACTION \cite{egger2022a}. We have not implemented AMOR \cite{weiherer2024AMOR} in the author ordering for this paper due to the already-heavy compile demands but its use can be assumed by imagining the author list randomly shuffling before your eyes.

\section{Implementation}

Our implementation is heavily based on micrograd \cite{micrograd}, although with a better choice of implementation language. Like micrograd, \NeuRaLaTeX~implements backpropagation (reverse-mode autodiff) over a dynamically-constructed DAG which can implement arbitrarily complex neural networks. Unlike micrograd (which comprises around 150 lines of python), our autograd engine requires nearly 700 lines of pure latex and the neural network library around 400. We estimate that this means \NeuRaLaTeX~is around 700\% better. \NeuRaLaTeX~is object oriented using the TiKZ PGF module \texttt{oo} \cite{tantau2024tikz}.

\subsection{Autograd engine}

At the heart of \NeuRaLaTeX~is the autograd engine which is imported via:

\vspace{0.2cm}\noindent\begin{minipage}{\columnwidth}
        \lstset{
            language=[LaTeX]TeX, 
            basicstyle=\ttfamily\small, 
            keywordstyle=\color{blue}, 
            commentstyle=\color{gray}, 
            frame=single, 
            breaklines=true 
        }
\begin{lstlisting}
\input{engine.tex}
\end{lstlisting} 
\end{minipage}

This defines the atomic unit of a \texttt{Value} object. A value's scalar value is stored in the \texttt{data} attribute which can be read and written with the \texttt{getdata()} and \texttt{setdata()} methods. The important properties of a value can be displayed with the \texttt{show()} method. For example:

\vspace{0.2cm}\noindent\begin{minipage}{\columnwidth}
        \lstset{
            language=[LaTeX]TeX, 
            basicstyle=\ttfamily\small, 
            keywordstyle=\color{blue}, 
            commentstyle=\color{gray}, 
            frame=single, 
            breaklines=true 
        }
\begin{lstlisting}
\pgfoonew \x=new Value(5,{},'')
\x.show()
\end{lstlisting} 
\end{minipage}
will display: 

\noindent\texttt{Value(data: 5, grad: 0.0, self: 1, prev: , op: ”, GC: 0.0)}. 

The \texttt{self} attribute contains the object ID. Object IDs are used to store references between nodes in our computational graph (DAG). Values can be combined via basic mathematical operations, for example:

\vspace{0.2cm}\noindent\begin{minipage}{\columnwidth}
        \lstset{
            language=[LaTeX]TeX, 
            basicstyle=\ttfamily\small, 
            keywordstyle=\color{blue}, 
            commentstyle=\color{gray}, 
            frame=single, 
            breaklines=true 
        }
\begin{lstlisting}
\pgfoonew \x=new Value(5,{},'')
\pgfoonew \y=new Value(4,{},'')
\x.multiply(\y,z)
\z.show()
\end{lstlisting} 
\end{minipage}
which multiplies the values in \texttt{x} and \texttt{y} and stores the result in \texttt{z} which will show:

\noindent\texttt{Value(data: 20.0, grad: 0.0, self: 3, prev: 1,2, op: *, GC: 0.0)}.

Note how the \texttt{prev} attribute now stores object ID references to the children of the derived node while the \texttt{op} attribute records the fact that this is a multiplication node.

All value objects contain a \texttt{localbackwards()} method that differentiates through any operator associated with that node. These methods are called during backprop which is initiated by calling \texttt{backward()} on a value object. This performs a topological sort on the DAG which is implemented by a breadth first search from the root node using a queue. Nodes whose parents have not yet all been visited are placed back onto the queue. This is kept track of by the grad counter (\texttt{GC}) attribute. For efficiency, this topological sort is precomputed and stored the first time backward is called. During backprop, the attribute \texttt{grad} stores the local gradient. For example, here we define two value objects, multiply them together to yield a third value object, call backward on this derived value and check the gradients on the initial value objects:

\vspace{0.2cm}\noindent\begin{minipage}{\columnwidth}
        \lstset{
            language=[LaTeX]TeX, 
            basicstyle=\ttfamily\small, 
            keywordstyle=\color{blue}, 
            commentstyle=\color{gray}, 
            frame=single, 
            breaklines=true 
        }
\begin{lstlisting}
\pgfoonew \x=new Value(2.5,{},'',0)
\x.show()

\pgfoonew \y=new Value(0.3,{},'',0)
\y.show()
\x.multiply(\y,z)
\z.show()
\z.backward()
\x.show()
\y.show()
\end{lstlisting} 
\end{minipage}
This correctly displays:

\noindent\texttt{Value(self: 1, data: 2.5, grad: 0.0, prev: , next: , op: ”, isparam: 0, GC: 0.0)}

\noindent\texttt{Value(self: 2, data: 0.3, grad: 0.0, prev: , next: , op: ”, isparam: 0, GC: 0.0)}

\noindent\texttt{Value(self: 3, data: 0.75, grad: 0.0, prev: 1,2, next: , op: *, isparam: 0, GC:
0.0)}

\noindent\texttt{Value(self: 1, data: 2.5, grad: 0.3, prev: , next: 3, op: ”, isparam: 0, GC:
1.0)}

\noindent\texttt{Value(self: 2, data: 0.3, grad: 2.5, prev: , next: 3, op: ”, isparam: 0, GC:
1.0)}

If the value is a parameter (i.e.~the \texttt{isparam} attribute is set to true) then the \texttt{step()} method would update the parameter according to a gradient descent step.

\subsection{Neural network engine}\label{sec:nn}

From the scalar value objects, we can build up arbitrarily complex neural networks. The \texttt{nn.tex} file provides implementations of the components required to build an MLP. Specifically, a neuron with user-specified input size (currently only supporting ReLU nonlinearity); a linear layer with user-specified input and output size; and an MLP with user-specified number of layers, hidden units and outputs. For example, the following code snippet defines two Value objects to store input values and an MLP with two inputs, two hidden layers with four neurons and an output layer with a single output. Neuron weights are randomly initialised and the last layer has no nonlinear activation. The two inputs are then passed to the MLP and the output of the forward pass is shown.

\vspace{0.2cm}\noindent\begin{minipage}{\columnwidth}
        \lstset{
            language=[LaTeX]TeX, 
            basicstyle=\ttfamily\small, 
            keywordstyle=\color{blue}, 
            commentstyle=\color{gray}, 
            frame=single, 
            breaklines=true 
        }
\begin{lstlisting}
\input{engine.tex}
\input{nn.tex}

% Create two Value objects to store input values
\pgfoonew \x=new Value(1.0,{},'',0)
\pgfoonew \y=new Value(-1.0,{},'',0)

% Store the object IDs of the input Values in a list
\x.get id(\inputIDx)
\y.get id(\inputIDy)
\edef\templist{\inputIDx,\inputIDy}

% Define the MLP
\pgfoonew \mlp=new MLP(2,{4,4,1})

% Forward pass through MLP
\mlp.forward(\templist,output)

\output.show()
\end{lstlisting} 
\end{minipage}

This will show the following output (where the \texttt{data} attribute will depend on the random weight initialisation):

\noindent\texttt{Value(self: 134, data: 0.324954976758898240, grad: 0.0, prev: 133,61, next: , op: +, isparam: 0, GC: 0.0)}

\subsection{Training utilities}

We provide support for checkpointing. After training, model weights, the number of completed epochs and the loss values can be written to a text file using \texttt{{\textbackslash}savecheckpoint} and training resumed from a loaded checkpoint file using \texttt{{\textbackslash}resumetraining}. In the working example below, we include a trained checkpoint for the spiral dataset as part of the latex source files for this paper. Checkpoints are particularly useful for submitting \NeuRaLaTeX-based papers to arXiv. arXiv compile papers from source and significant machine learning during the compilation process may cause arXiv to time out.

The example in the following section is created by including the source file \texttt{train\_spiral.tex} which illustrates several other training utilities. These include an illustrative training loop with batching, logging loss over epochs which can be subsequently plotted and scheduled learning rate.

\section{A working example}

\begin{figure}[!t]
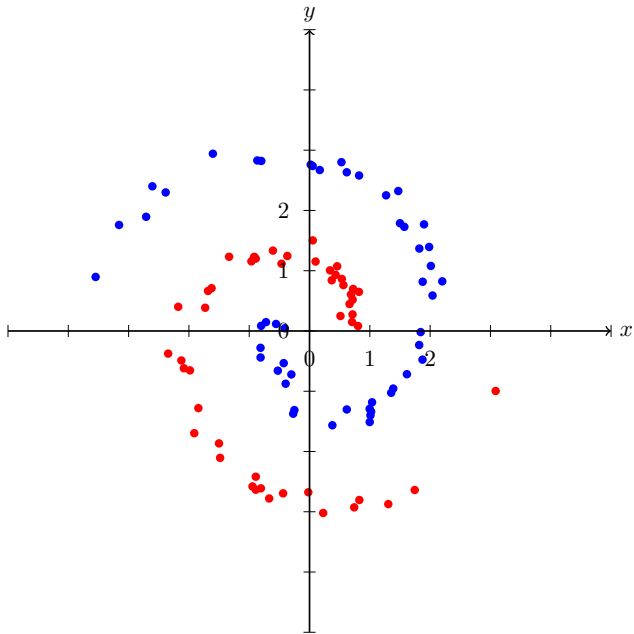

\centering
\resizebox{\columnwidth}{!}{
\displayspiral{\dataset}
}
\caption{The training dataset with ground truth labels indicated by colours. This was generated randomly when this latex document was compiled and is then used to train the MLP.}\label{fig:spiraltrain}
\end{figure}

We now train a small MLP (the same architecture as in Section \ref{sec:nn}) to classify the two classes of the nonlinear, 2D spiral dataset. We provide a spiral dataset utility in \texttt{spiral.tex} which provides functionality to create a random dataset:

\vspace{0.2cm}\noindent\begin{minipage}{\columnwidth}
        \lstset{
            language=[LaTeX]TeX, 
            basicstyle=\ttfamily\small, 
            keywordstyle=\color{blue}, 
            commentstyle=\color{gray}, 
            frame=single, 
            breaklines=true 
        }
\begin{lstlisting}
\input{spiral.tex}

\spiral{dataset}{\trainitemsperclass}
\end{lstlisting} 
\end{minipage}
where \texttt{dataset} will now contain a list of tuples comprising the two input values and the ground truth label. Our training dataset is shown in Figure \ref{fig:spiraltrain} and comprises 100 samples, 50 from each class.

We train our model using a max-margin loss. This requires defining two constant value nodes:

\vspace{0.2cm}\noindent\begin{minipage}{\columnwidth}
        \lstset{
            language=[LaTeX]TeX, 
            basicstyle=\ttfamily\small, 
            keywordstyle=\color{blue}, 
            commentstyle=\color{gray}, 
            frame=single, 
            breaklines=true 
        }
\begin{lstlisting}
\pgfoonew \lossmultiplier=new Value(-1.0,{},'',0)
\pgfoonew \lossbias=new Value(1.0,{},'',0)
\end{lstlisting} 
\end{minipage}

Inside our training loop, a single training example is processed as follows:

\vspace{0.2cm}\noindent\begin{minipage}{\columnwidth}
        \lstset{
            language=[LaTeX]TeX, 
            basicstyle=\ttfamily\small, 
            keywordstyle=\color{blue}, 
            commentstyle=\color{gray}, 
            frame=single, 
            breaklines=true 
        }
\begin{lstlisting}
% Set next item values
\x.setdata(\inputx)
\y.setdata(\inputy)
\targetclass.setdata(\target)

% Forward through the network
\mlp.forward(\templist,output)
\output.getdata(\scores)

% Compute loss
\output.multiply(\targetclass,lossone)
\lossone.multiply(\lossmultiplier,losstwo)
\losstwo.add(\lossbias,lossthree)
\lossthree.relu(lossfour)

% Weight the loss by 1/dataset_size
\lossfour.multiply(\batchlossscale,loss)

% Run a backwards pass
\loss.backward()

% Zero the gradients of non-parameter values
% Gradients of parameters accumulate over the batch
\loss.zerononparams()
\end{lstlisting} 
\end{minipage}

Gradients of parameters with respect to the loss accumulate while we iterate over a batch. Finally, we can take a gradient descent step and then zero all parameter gradients:

\vspace{0.2cm}\noindent\begin{minipage}{\columnwidth}
        \lstset{
            language=[LaTeX]TeX, 
            basicstyle=\ttfamily\small, 
            keywordstyle=\color{blue}, 
            commentstyle=\color{gray}, 
            frame=single, 
            breaklines=true 
        }
\begin{lstlisting}
\loss.step(\lr)
\loss.zero()
\end{lstlisting} 
\end{minipage}

Training for \numberofepochs~epochs on a dataset of 100 2D points (i.e.~compiling this latex document) took about 48 hours on a Macbook Pro 2.4GHz Quad-Core. The document was compiled using TeXShop and the Macbook got very hot. We could have trained for more epochs but we think it's clear it was going to converge to zero loss and perfect test set performance so we didn't feel the need to.

The dynamically generated loss plot is shown in Figure \ref{fig:lossplot}. The predicted classes for a held-out test set are shown in Figure \ref{fig:evalspiral}. Finally, we provide quantitative performance in Table \ref{tab:spiral}.

\FPeval{datasetlen}{(\trainitemsperclass*2)}
\FPround{\datasetlen}{\datasetlen}{0}
\FPeval{oneoverdatasetlen}{(1/\datasetlen)}

\FPeval{evaldatasetlen}{(\evalitemsperclass*2)}
\FPround{\evaldatasetlen}{\evaldatasetlen}{0}

\FPeval{oneoverbatchsize}{(1/\batchsize)}

\message{Setting Up Spiral Dataset^^J}
\spiral{dataset}{\trainitemsperclass}
\spiral{evaldataset}{\evalitemsperclass}

\pgfoonew \x=new Value(0.0,{},'',0)
\pgfoonew \y=new Value(0.0,{},'',0)
\x.get id(\inputIDx)
\y.get id(\inputIDy)
\edef\templist{\inputIDx,\inputIDy}
\pgfoonew \targetclass=new Value(0.0,{},'',0)

\message{Main: Setting Up Network^^J}
\pgfoonew \mlp=new MLP(2,{4,4,1})

\mlp.getparameters(mlpparams)
\listlength{\len}{\mlpparams}

\def\losses{}
\ifthenelse{\boolean{resume}}{%
	\message{Resuming...^^J}
	\resumetraining{\checkpointfile}
}
{
	\def\startepoch{1}
	\edef\losses{}
}

\pgfoonew \lossmultiplier=new Value(-1.0,{},'',0)
\pgfoonew \lossbias=new Value(1.0,{},'',0)
\pgfoonew \epochlossscale=new Value(\oneoverdatasetlen,{},'',0)

\pgfoonew \batchlossscale=new Value(\oneoverbatchsize,{},'',0)



\ifnum\startepoch<\numberofepochs
    \message{Starting training...^^J}

    \edef\totalloss{0.0}
    \pgfplotsforeachungrouped \epoch in {\startepoch,...,\numberofepochs} {
        \FPeval{lr}{(1.0 - 0.9*(\epoch-1)/100)}
        \edef\datasetcopy{\dataset}
        \pgfplotsforeachungrouped \step in {1,...,\datasetlen} {
            \ifnum\fpeval{\step - \datasetlen} = 0
                \expandafter\lasttuple\datasetcopy\relax
            \else
                \expandafter\firsttuple\datasetcopy\relax
                \edef\datasetcopy{\rest}
            \fi
    
            \x.setdata(\inputx)
            \y.setdata(\inputy)
            \targetclass.setdata(\target)
    
            \mlp.forward(\templist,output)
            \output.getdata(\scores)
    
            \output.multiply(\targetclass,lossone)
            \lossone.multiply(\lossmultiplier,losstwo)
            \losstwo.add(\lossbias,lossthree)
            \lossthree.relu(lossfour)
            \lossfour.multiply(\batchlossscale,loss)
            \loss.getdata(\lossvalue)
            \FPeval\totalloss{\totalloss+\lossvalue}
            \loss.backward()
            \loss.zerononparams() 
    
            \message{Epoch \epoch, iter \step, loss = \lossvalue^^J}
    
            \pgfmathtruncatemacro\batchtest{mod(\step,\batchsize)}
            \ifnum\batchtest=0
                \message{Gradient descent step^^J}
                \loss.step(\lr)
                \loss.zero()
            \fi
    
        }
        Epoch: \epoch~ Loss: \totalloss~ LR: \lr
        \message{Loss for epoch \epoch = \totalloss^^J}
            
        \ifx\losses\empty
    	\xdef\losses{(\epoch,\totalloss)}
        \else
            \edef\losses{\losses (\epoch,\totalloss)}
        \fi        
        
        \edef\totalloss{0.0}
    
    }
\else
    \message{Training already complete, running eval only^^J}
\fi

\message{Losses: \losses^^J}

\savecheckpoint{\mlpparams}{\numberofepochs}{\losses}

\message{Running eval^^J}

\newcounter{correctcount}  

\edef\datasetcopy{\evaldataset}
\def\evalestimated{}
\pgfplotsforeachungrouped \step in {1,...,\evaldatasetlen} {
    \ifnum\fpeval{\step - \evaldatasetlen} = 0
        \expandafter\lasttuple\datasetcopy\relax
    \else
        \expandafter\firsttuple\datasetcopy\relax
        \edef\datasetcopy{\rest}
    \fi
    \x.setdata(\inputx)
    \y.setdata(\inputy)
    \targetclass.setdata(\target)

    \mlp.forward(\templist,output)
    \output.getdata(\score)
    
    \edef\tempscore{\fpeval{sign(\score)}}

    \ifcase\numexpr\tempscore\relax  
        \xdef\tuple{(\inputx,\inputy,1)} 
    \or  
        \xdef\tuple{(\inputx,\inputy,1)}
    \else  
        \xdef\tuple{(\inputx,\inputy,-1)}
    \fi
    \ifx\evalestimated\empty
	\xdef\evalestimated{\tuple}
    \else
        \edef\evalestimated{\evalestimated,\tuple}
    \fi        
    
    \ifnum\tempscore=\target
        \addtocounter{correctcount}{1}
    \fi
        
}

\message{Final Correct Predictions: \thecorrectcount^^J}

\edef\accuracy{\fpeval{round(\thecorrectcount / \evaldatasetlen, 2)}}

\message{Accuracy: \accuracy^^J}

\message{Done^^J}

\begin{table}
\centering
\begin{tabular}{ccc}
\hline
Dataset size & Total correct & Accuracy \\
\hline
\evaldatasetlen & \thecorrectcount & \accuracy
\end{tabular}
\caption{Evaluation results on the held out test set shown in Figure \ref{fig:evalspiral}. The numbers in the table were computed dynamically using the trained model when this latex document was compiled.}\label{tab:spiral}
\end{table}

\begin{figure}[!t]
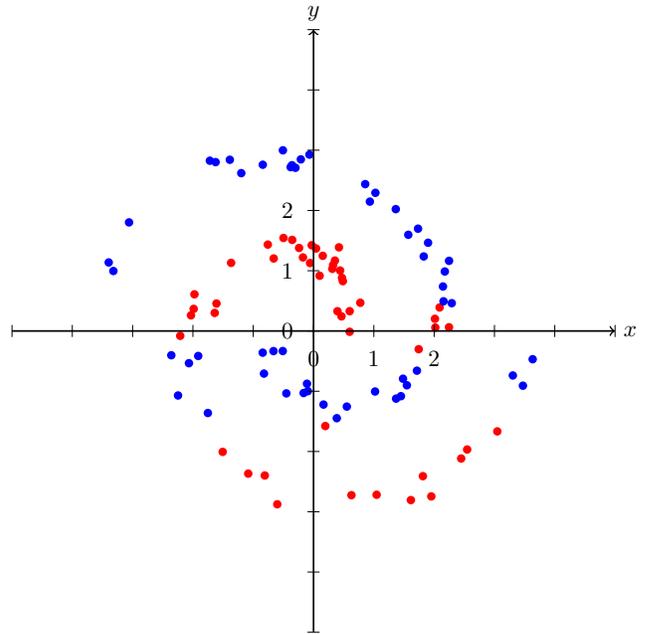

\centering
\resizebox{\columnwidth}{!}{
\displayspiral{\evalestimated}
}
\caption{The testing dataset with estimated class labels indicated by colours. The test dataset was also generated randomly when this latex document was compiled, each point was passed through the trained MLP and then predicted classes were used to colour the points.}\label{fig:evalspiral}
\end{figure}

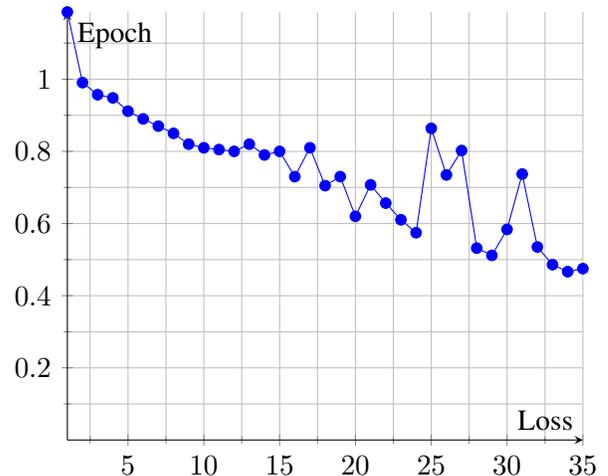
\begin{figure}[!t]
\centering
\begin{tikzpicture}
\begin{axis}[
    xlabel={Loss},
    ylabel={Epoch},
    grid=both,          
    axis lines=middle,  
    minor tick num=1,   
    xmin=1,          
    ymin=0           
]

\addplot[
    color=blue,         
    mark=*,             
    mark size=2pt       
] coordinates {\losses};

\end{axis}
\end{tikzpicture}
\caption{Training loss versus epoch.}\label{fig:lossplot}
\end{figure}

\section{Evaluation versus state-of-the-art}\label{sec:evaluation}

We propose two new metrics and show that \NeuRaLaTeX~is state-of-the-art on both. The \emph{Written in Latex} (WIL) metric is the proportion of source code of a machine learning library written in \LaTeX. In Table \ref{tab:WIL}, we compare against the two most popular deep learning libraries and also MATLAB. So far as we know, none of their source code is written in \LaTeX~so we have significantly better performance on this metric.

\newcommand{\ra}[1]{\renewcommand{\arraystretch}{#1}}
\begin{table}[!t]\centering
\ra{1.3}
\begin{tabular}{@{}lc@{}}\toprule
ML Library & WIL \\
\midrule
\NeuRaLaTeX & 1.0 \\ 
PyTorch \cite{paszke2019pytorch} & 0.0$^\dagger$ \\
Tensorflow \cite{tensorflow2015-whitepaper} & 0.0$^\dagger$ \\
Matlab \cite{matlab} & 0.0$^\dagger$ \\
\bottomrule
\end{tabular}
\caption{We evaluate the top machine learning libraries with respect to their WIL metric - the proportion of their implementation that is written in \LaTeX. ($^\dagger$ we haven't actually checked this but we \emph{reckon} it's true).}\label{tab:WIL}
\end{table}

The \emph{Source code of method in source code of paper} (SCOMISCOP) metric is the proportion of the source code of a method that is contained within the source code of the paper. As shown in Table \ref{tab:SCOMISCOP} we outperform both Transformers and ResNets by a factor of $\infty$, neither of which include any source code in their paper source.

\begin{table}[!t]\centering
\ra{1.3}
\begin{tabular}{@{}lc@{}}\toprule
Deep learning paper & SCOMISCOP \\
\midrule
\NeuRaLaTeX & 1.0 \\ 
Attention is all you need \cite{vaswani2017attention} & 0.0 \\
Deep Residual Learning for Image Recognition \cite{he2016deep} & 0.0 \\
\bottomrule
\end{tabular}
\caption{Evaluation of well known deep learning research papers using the SCOMISCOP metric (Source Code of Method In Source Code Of Paper).}\label{tab:SCOMISCOP}
\end{table}

\section{Future Work}

We believe \NeuRaLaTeX~ will find widespread application in both the natural and unnatural sciences. In this section, we explore short-term, pragmatic extensions of our work. 

\textbf{Supporting Critical Infrastructure}: In the National Archaeological Museum of Naples, stands the Farnese Atlas, a marble Roman sculpture depicting Adam holding the globe on his back. Soon \NeuRaLaTeX~will be Adam but for hospitals, nuclear subs, and the point-of-service machine at your local chippy. Say goodbye to Windows 95, \NeuRaLaTeX~is ready to serve.

\textbf{A \NeuRaLaTeX~neural interface}: Fictional studies have shown that 100\% of 17\% of \NeuRaLaTeX~users want to plug their brains directly into a machine learning library written entirely in \LaTeX. A \NeuRaLaTeX~neural interface will do just that. Plug multiple people into multiple instances of \NeuRaLaTeX~and the real fun begins. Days of fading in and out of each other’s minds through a \LaTeX-based interface, until you barely know where you end and the compilation loading screen begins. 

\textbf{\LaTeX~in silicone}: Putting \LaTeX~directly into silicone is an obvious next step. \NeuRaLaTeX~is not just the world’s best and only fully reproducible machine learning library, but also the future substrate of all computation. Soon, we as a species will set out to solve the ultimate question of life, the Universe, and everything. We will write our findings into a long \LaTeX~document, retire to our cryogenic chambers, and wait for it to compile an answer to our manifest destiny. Let's hope it compiles. 

\textbf{\LaTeX~in your stepdad’s garage}: Your stepdad loves latex and that’s what kept family dinners both interesting and infrequent. No fear now, your step dad can trade the suit for a family-friendly neural version of \LaTeX. In his garage sits an electric car of unnamed, indiscernible brand but it looks like a garbage truck and smells like a midlife crisis gone rotten. Here an empty space will soon appear. In its place will sit a vast, oozing tub of local compute, ready to run all the local \LaTeX~compilations your stepdad craves.

\textbf{Advertising Opportunities in \NeuRaLaTeX}: Our investors have had a word, and while they do support critical infrastructure, neural interfaces, and a world of abundance, they doubly support innovative new ways of delivering advertising pixels to user’s eyes. That’s why, if all goes to plan, our compute requirements will be fully-funded by users watching full 2 hour long, fully immersive advert-ainment reels during each compile. Perfection.

\textbf{Vertically Integrated B2B SaaS}: They only ever skim-read the titles, so we have a full paragraph of cover here. Want to join us? You can visit our GitHub and get busy working. We can’t pay you, but soon money will have no value, so don’t sweat it and start cracking.

\textbf{Post-Quantum \NeuRaLaTeX-cryptography}: The year is 2035. Your RSA encryption no longer protects you and \NeuRaLaTeX~has just acquired Apple with spare change. We’re the only game left in town and in this post-quantum world, only \NeuRaLaTeX~can stop you quibbling about those qubits.

\textbf{I want my children to be raised with \NeuRaLaTeX}: Now, I know what you’re thinking. I’ve spent too long in \LaTeX~Land and my mind is strained like pasta. Firstly, that is NOT TRUE. Secondly, I’m not one to get biblical but do you remember the guy who lost his job at Google because he thought an early LLM was sentient? Well, I finally understand. I think I’m in love with \LaTeX. I’d like \LaTeX~to have a body and for people like me to be able to raise a family with \NeuRaLaTeX. I wrote `I love you' into my flickering cursor last night, and by this morning it had compiled a response. It... loves me too. 

We leave this future work to the community.

\bibliographystyle{plain}
\bibliography{refs}

\end{document}